\documentclass[twocolumn]{svjour3}          
\smartqed  
\usepackage{graphicx}
\usepackage[table,xcdraw,dvipsnames]{xcolor}
\usepackage{multirow}
\usepackage{array, booktabs, makecell}
\usepackage{hyphenat}
\usepackage{caption}
\usepackage{amsmath}
\usepackage{url}
\usepackage{tikz}
\usepackage[round]{natbib}
\usepackage[colorinlistoftodos,prependcaption,textsize=tiny]{todonotes}
\usepackage{xifthen}
\usepackage[linesnumbered,ruled,vlined]{algorithm2e}
\usepackage[caption=false]{subfig}
\usepackage{pgfplots, pgfplotstable,booktabs}
\usetikzlibrary{positioning,patterns}
\graphicspath{ {./img/} }

\definecolor{dslmauve}{RGB}{142,46,121}
\definecolor{dslblue}{RGB}{100,152,176}
\definecolor{dsldarkyellow}{RGB}{171,166,26}
\definecolor{dslgreen}{RGB}{23,107,23}
\definecolor{dslgrey}{RGB}{145,145,145}

\definecolor{true}{RGB}{0,150,0}
\definecolor{false}{RGB}{220,0,0}
\definecolor{dead}{RGB}{192,192,192}

\newcommand{\fabnote}[2][]{\ifthenelse{\isempty{#1}{}}{
\todo[inline,backgroundcolor=LimeGreen]{\textbf{FPE:} #2}
}{
\todo[inline,backgroundcolor=LimeGreen,#1]{\textbf{FPE:} #2}}
}
\newcommand{\brunote}[2][]{\ifthenelse{\isempty{#1}{}}{
\todo[inline,backgroundcolor=Thistle]{\textbf{BLE:} #2}
}{
\todo[inline,backgroundcolor=Thistle,#1]{\textbf{BLE:} #2}}
}
\newcommand{\aymnote}[2][]{\ifthenelse{\isempty{#1}{}}{
\todo[inline,backgroundcolor=SkyBlue]{\textbf{ACR:} #2}
}{
\todo[inline,backgroundcolor=SkyBlue,#1]{\textbf{ACR:} #2}}
}
\newcommand{\alexnote}[2][]{\ifthenelse{\isempty{#1}{}}{
\todo[inline,backgroundcolor=BurntOrange]{\textbf{AVE:} #2}
}{
\todo[inline,backgroundcolor=BurntOrange,#1]{\textbf{AVE:} #2}}
}
\newcommand{\antnote}[2][]{\ifthenelse{\isempty{#1}{}}{
\todo[inline,backgroundcolor=Purple]{\textbf{ACH:} #2}
}{
\todo[inline,backgroundcolor=Purple,#1]{\textbf{ACH:} #2}}
}

\newcommand*\circled[1]{\tikz[baseline=(char.base)]{
           \node[shape=circle,draw,inner sep=1pt] (char) {#1};}}
           
\begin{document}

\title{DAE : Discriminatory Auto-Encoder for  multivariate time-series anomaly detection in air transportation}
\author{Antoine Chevrot \and
        Alexandre Vernotte         \and
        Bruno Legeard
}

\institute{A. Chevrot, A. Vernotte, B. Legeard \at
              DISC/FEMTO-ST Institute, UBFC, CNRS \\
              Besan\c con, France \\
              Tel.: +333-81-66-20-87\\
              \email{name.lastname@femto-st.fr}
}

\date{Received: date / Accepted: date}

\maketitle

\begin{abstract}

The Automatic Dependent Surveillance\hyp{}Broadcast protocol is one of the latest compulsory advances in air surveillance. 
While it supports the tracking of the ever-growing number of aircraft in
the air, it also introduces cybersecurity issues that must be mitigated e.g., false data injection attacks where an attacker emits fake surveillance information. 
The recent data sources and tools available to obtain flight tracking
records allow the researchers to create datasets and develop Machine Learning models capable of detecting such anomalies in En-Route trajectories.  
In this context, we propose a novel multivariate anomaly detection model called Discriminatory Auto-Encoder (DAE). It uses the baseline of a regular LSTM-based auto-encoder but with several decoders, each getting data of a specific flight phase (e.g. climbing, cruising or descending) during its training. 
To illustrate the DAE's efficiency, an evaluation dataset was created using real-life anomalies as well as realistically crafted ones, with which the DAE as well as three anomaly detection models from the literature were evaluated. 
Results show that the DAE achieves better results in both accuracy and speed of detection. The dataset, the models implementations and the evaluation results are available in an online repository, thereby enabling replicability and facilitating future experiments. 

\end{abstract}

\section{Introduction}
\label{sec:intro}

Over the past ten years, air traffic control has faced a growing number of users and the traffic load 
keeps growing steadily. With an increasingly congested airspace, numerous new 
issues are appearing such as flight delays. 
This increases the overall cost of the flights and exacerbates an already existing tendency for 
air companies to close down\footnote{https://www.nbcnews.com/news/world/british-travel-firm-thomas-cook-collapses-stranding-hundreds-thousands-n1057456} in favour of low-cost companies. In another vein, congestioned 
airports imply that the planes stay longer in taxiways which is where they are the 
least efficient~\citep{Polishchuk2019}, increasing their fuel consumption as well as their particle emissions~\citep{Zhang2019}.

\indent To tackle these new challenges, Air Traffic Control (ATC) needs improved 
surveillance technologies supporting the constraints in terms of simultaneously 
handled aircraft as well as overall accuracy. The Automatic Dependent Surveillance-Broadcast 
(ADS-B) protocol is currently being deployed world-wide in an effort to improve flights 
management. ADS-B requires participating aircraft to broadcast their information periodically 
in an encoded message, like a beacon. 
\indent This technology embodies the shift from independent and non-cooperative 
surveillance technologies, historically used for aircraft surveillance, to dependent 
and cooperative technologies. In this new context, ground stations need aircraft to 
cooperate and are dependent on aircraft's Global Navigation Satellite System (GNSS) 
capabilities to determine their position. 

\indent Nonetheless, ADS-B is not a new protocol. The ICAO (International Civil Aviation 
Organization) issued a plan in 2002 \footnote{\url{https://www.icao.int/publications/Documents/9750_2ed_en.pdf}} 
recognizing ADS-B as an emerging technology for dissemination of aircraft position 
information. 
In 2021, ADS-B is now compulsory in most air-spaces but the protocol 
itself stayed sensibly the same as it was imagined twenty years ago and the security was 
not in the highest priority. As a result, anyone with the proper equipment can receive and create messages freely.
This liberty in both emission and 
reception make ADS-B vulnerable to spoofing, and more precisely to attacks called 
FDIA --- False Data Injection Attack --- which purpose is to create fake surveillance 
messages respecting conscientiously the protocol in order to dupe the air traffic 
controllers to believe in an abnormal situation. \\

\indent Although ADS-B is not the only protocol used for flights tracking 
-- e.g radar technologies --, it is, as of today, a central brick in the means 
of surveillance used by public air transportation. In this context, there has been a 
growing interest for conducting research on anomaly detection systems that address 
these new threats~\citep{Strohmeier2015b}. Among the different existing solutions,
some are based on Machine Learning (ML) anomaly detection models. These models already 
find applications in many different domains like power systems~\citep{Wang2018} or
sensor networks~\citep{Malhotra2016} and are found quite popular in recent years.
One downside of these models is their need for consequent data availability to achieve meaningful results.
It is indeed critical for ML researchers to have access to reliable and genuine data sources to train their models. 
Thankfully, for ADS-B data, the OpenSky Network~\citep{Schafer2014} is one of the references in terms of accessibility
and data history in air transportation, and one can easily obtain surveillance data from almost anywhere on the globe.
This access to genuine data and the lacks of anomalous ones in comparison favours one particular architecture of ML model called auto-encoders.

\indent Auto-encoders are unsupervised ML models often used for anomaly detection and can be found in the literature in many different forms. These models use a first network called the encoder which \emph{encodes} the input data into a latent representation which is then \emph{decoded} by a second network called the decoder. 
The discrepancies between the input data and the output
ones are then used to detect anomalies in the original data. They can be coupled with Recurrent Neural Networks -
RNN - to address the temporality of the data~\citep{Malhotra2016}. Shown to be quite effective, they have already been
used in the past to detect different types of anomalies in the ADS-B protocol like en-route trajectory anomalies
\citep{Olive2019} or spoofing attempts~\citep{Ying2019}.

\indent This paper presents a novel type of auto-encoder to use for anomaly detection in ADS-B. The main contributions of this work is listed hereafter:

\begin{itemize}
\item[(i)] \emph{The DAE} --Discriminatory auto-encoder--, a novel asymetric auto-encoder addressing fluctuations in time series. 
To the best of our knowledge, this is the first time auto-encoders are used with a single encoder connected to several decoders for anomaly detection in time-series. 
\item[(ii)] \emph{The full data framework} using existing tools includes the data cleaning, the feature extraction and the data serialization for model training. Emphasis is made on replicability through a code repository publicly accessible.
\item[(iii)] \emph{Realistic and replicable validation scenarios} are created using an alteration tool to experiment with different types of anomalies. It results in a dataset also available online to compare future models and provide a common base for benchmarks and studies.
\item[(iii)] \emph{Experimental results} using the abovementioned validation scenarios to compare the different existing solutions of ML anomaly detection showing that the DAE performs well overall.
\end{itemize}

To present the model and the different results achieved with it, this paper has been organized as follows: 

\indent \emph{Section~\ref{sec:bg}} provides a basis for understanding the ADS-B protocol, an explanation on FDIAs and the risks associated with it. \emph{Section~\ref{sec:rw}} presents previous works done on anomaly detection for the ADS-B with an emphasis on Machine Learning based techniques. \emph{Section~\ref{sec:dae}} introduces the novel anomaly detection model developed in this paper by detailing its architecture. Section~\ref{sec:data} details the process of data gathering and processing to obtain proper training data for the model. \emph{Section~\ref{sec:eval}} presents the evaluation of this paper, showcasing the data used and the different results obtained using different anomaly detection models. Follow some discussions about implementation and caveats in \emph{Section~\ref{sec:disc}}.  \emph{Section~\ref{sec:conc}} concludes this paper.

\section{Background}
\label{sec:bg}

\subsection{ADS-B overview}
Communication via ADS-B consists of aircraft using a Global Navigation Satellite 
System (GNSS) to determine their position and broadcasting it periodically without 
solicitation (a.k.a beacons or squitters), along with other information obtained 
from on-board systems such as altitude, ground speed, aircraft identity, heading, etc.
Ground stations pick up on the squitters, process them and send the information out 
to the ATC system. 
The ADS-B data link is generally carried on the 1090MHz frequency. 
ADS-B is therefore a cooperative (aircraft need a transponder) and dependent 
(on aircraft data) surveillance technology, which constitutes a fundamental 
change in ATC. It means for instance that not only ground stations with antennas 
positioned at the right angle and direction can receive position information. 
Aircraft can now receive squitters from other aircraft, which notably improves cockpit 
situational awareness as well as collision avoidance. \\
\indent The introduction of ADS-B also provides controllers with improved situational 
awareness of aircraft positions in En-Route and TMA (Terminal Control Area) airspaces, 
and especially in NRAs (Non Radar Areas). It theoretically gives the possibility of 
applying much smaller separation minima (e.g., from 80NM longitudinal separation to 
just 5~NM in NRAs) than what is presently used with current procedures 
(Procedural Separation)~\citep{EUROC05}.
It has a much greater accuracy and update rate, with a smaller latency. The major 
drawback of the technology lies in its lack of encryption and authentication, which 
is discussed in the following section.

\subsection{False Data Injection Attacks}

 The progressive shift from independent and non\hyp{}cooperative technologies 
 (PSR/SSR \citep{skolnik1970radar}) to dependent and cooperative technologies 
 (ADS-B) has created a strong reliance on external entities (aircraft, GNSS) 
 to estimate aircraft state. This reliance, along with the introduction of 
 air-to-ground data links via Modes A/C/S and the broadcast nature of ADS-B, 
 has brought alarming cyber security issues. Extensive research can be found in 
 the literature that discuss these issues~\citep{SchaeLM13,ZhangLLN17,WessoHE14,StrohSPLM16}, 
 stressing that the introduction of ADS-B has enabled a class of attack referred to 
 as \emph{False Data Injection Attacks} (FDIAs).

FDIAs were initially introduced in the domain of wireless sensor networks~\citep{Ma08}. 
A wireless sensor network is composed of a set of nodes (i.e. sensors) that send data 
report to one or several ground stations. Ground stations process the reports to reach 
a consensus about the current state of the monitored system. A typical scenario consists 
of an attacker who first penetrates the sensor network, usually by compromising one or 
several nodes, and thereafter injects false data reports to be delivered to the base 
stations. This can lead to the production of false alarms, the waste of valuable network 
resources, or even physical damage. Active research regarding FDIAs has been conducted in 
the power sector, 
mainly against smart grid state estimators~\citep{DanS10,LiuNR11}. It shows that these 
attacks may lead to power blackouts but can also disrupt electricity markets~\citep{XieMS10}, 
despite several integrity checks. 

FDIAs also exist in the domain of air traffic surveillance. 
Because surveillance relies on the information provided by aircraft's transponders 
to ground stations, aircraft transponders are equivalent to nodes from a wireless 
network, and ground stations are equivalent to base stations. Although in the ATC 
domain, there is no real effort to penetrate the sensor network, as all communications 
are unauthenticated and in clear text.  
Still, performing FDIAs on surveillance communications requires a deep understanding 
of the system, its protocol(s) and its logic, to covertly alter the surveillance flow. 
These attacks are much more complex to achieve than e.g., jamming, because the logic 
of the communication flow must be preserved and the falsified data must appear probable.


The means of the attacker to conduct FDIAs against ADS-B communications have already 
been detailed in previous work~\citep{Stroh16,ManesK17}. Considering the attacker has 
the necessary equipment, they can perform three malicious basic operations:
\begin{itemize}
\item[(i)] \emph{Message injection} which consists of emitting non-legitimate but 
well-formed ADS-B messages. 
\item[(ii)] \emph{Message deletion} which consists of physically deleting targeted 
legitimate messages using destructive or constructive interference. It should be 
noted that message deletion may not be mistaken for jamming, as jamming blocks all 
communications whereas message deletion drops selected messages only.
\item[(iii)] \emph{Message modification} which consists of modifying targeted 
legitimate messages using overshadowing, bit-flipping or combinations of message 
deletion and message injection. \\
\end{itemize}
One can sense the potential for disaster if one of these operations 
was to be executed successfully. 
It is of the utmost importance that none of the scenarios represent a 
real threat to such a critical infrastructure with human lives on the line.
However, because of the inherent properties of the ADS-B protocol, current 
solutions for securing ADS-B communications are only partial or involve an 
unbearable cost~\cite{StrohSPLM16}. Therefore, ATC systems must become robust 
against FDIAs, i.e. being capable of automatically detecting any tempering 
with the surveillance communication flow while being able to maintain the 
infrastructure in a working state.

\section{Related Work}
\label{sec:rw}

Several takes on improving the security of the ADS-B protocol can be found in the literature. These efforts often fall under several categories but for clarity's sake, these have been separated into two main categories here: one grouping technologies like multilateration or encryption to name a few and the other one grouping Machine Learning based techniques. 

\subsection{Security solutions for ADS-B protocol}

Many works on securing the ADS-B protocol using different technologies already exist with different degrees of feasibility. First, multilateration techniques or MLAT can be used to determine an aircraft position based on measures of time of arrivals (TOAs) of radio feeds. Each ADS-B message is timestamped and broadcast by aircraft. If several radars with synchronized clocks and known positions received the same ADS-B feed, then it is possible to calculate the position of the aircraft based on the differences of TOAs. MLAT can be used to detect ADS-B anomalies~\citep{Monteiro2015} and has the advantage to be very accurate. Recent work from~\citet{Zhao2020} also use MLAT to improve the ADS-B protocol accuracy as well as increasing the robustness of the surveillance systems. \citet{Fute2019} show experimentally that FDIAs can also be created to attack multilateration systems assuming an organized attacker with several devices to emit fake ADS-B proving that MLAT can have ultimately similar issues as ADS-B w.r.t. FDIAs.

\indent Regarding the use of the physical layer information,~\citet{Strohmeier2015a} create an intrusion detection system based on the strength of the signals they received from 2 different sensors. Similarly,~\citet{Schaefer2016} use the Doppler shift measurements to verify the En-Route positions of aircraft over time. Using the clocking system of the Mode-S sensors,~\citet{Leonardi2019} manages to obtain similar results to multilateration systems regarding on-board anomalies without the hassle of having at least 4 different sensors. \citet{Yang2019} used Machine Learning methods such as Gradient Boosting or Support Vector Machine to successfully flag anomalies on the PHY-layer features of ADS-B.

\indent On another level, several solutions were proposed for encrypting ADS-B. Based on the identity of aircraft,~\citet{Baek2017} describe a confidentiality framework to encrypt the ADS-B. Similarly,~\citet{Cook2015} uses Public Key Infrastructure (PKI) to try and secure ADS-B but it either suppresses the open characteristics of ADS-B or requires a change in the protocol itself. Further discussion and analysis can be found in a survey by~\citet{Strohmeier2015b}. 

\subsection{Machine Learning based anomaly detection techniques}

It is quite common for anomaly detection approaches to rely on multiple sources 
of data that each represents one aspect of the environment it characterizes, in 
order to find hidden/complex relationships between sources and rely on these to 
identify abnormal situations. As an example, one can use satellite 
images to check real positions of aircraft to confront them against received ADS-B data~\citep{Kastelic2019}.

In the air surveillance domain, there are multiple surveillance mechanisms 
(ADS-B, but also legacy radar technologies, multilateration, voice communication, etc) 
working simultaneously so that air traffic controllers get the clearest possible 
air situation picture. But there are many areas of the world where most mechanisms 
could not be deployed, and the sole automated technology that could is ADS-B 
(e.g., in the center of Australia, or anywhere offshore). 
Therefore, one of the goals of this work is to find out whether ADS-B alone 
is enough data to train machine learning models toward the detection of false data reports.

Multivariate time series anomaly detection is an active topic in the Machine Learning community. 
Supervised learning methods~\citep{Karam2020} require the training data to be labeled and thus can only identify anomalies found in said data. 
As a result, supervised models have limited usability here and unsupervised approaches are preferred.

Several efforts toward securing ADS-B using unsupervised Machine Learning techniques
can be found in the literature. \citet{Li2020} use an hidden Markov model to predict hidden states of the ADS-B
protocol and uses them to analyze the deviations during attacks to detect them. Its results however did not show any
advantages compared to other Machine Learning solutions like Recurrent Neural Network (RNN) based models in terms of
accuracy or false positive rate (FPR). Auto-encoders (AE) are a deterministic family of unsupervised machine learning anomaly detection
models often used in the latest publications concerning ADS-B security. 
Often coupled with RNNs like LSTM or GRU~\citep{Malhotra2016}, they
have shown good accuracy to detect coarse anomalies~\citep{Habler2018}, or more specific behaviours~\citep{Olive2018,Olive2019}.
\citet{Li2019} use LSTM-based auto-encoders in a generative adversarial network (GAN) as a mean to avoid an anomaly
threshold selection but misses proper metrics like recall or precision to correctly prove the efficiency of this method.
\citet{Akerman2019} use similar LSTM-AE along with convolutional networks to provide images of the traffic and the anomalies to improve
the user experience of such solution.

A stochastic variation of the auto-encoder called variational auto-encoder (VAE) are shown to be usable on detecting anomalies in
time-series like in the works of~\citet{Park2018}. Unlike a traditional auto-encoder, which maps the input onto a latent vector, a VAE maps the input
data into the parameters of a probability distribution, such as the mean and variance of a Gaussian. Applied to the anomaly detection
for ADS-B,~\citet{Luo2021} uses an LSTM-VAE model coupled to a support vector data description (SVDD) model to automatically generate its
anomaly threshold showing good results on similar coarse anomalies introduced by~\citet{Habler2018}. However, as pointed by
\citet{Su2019}, simply coupling LSTM and VAE together ignores the temporal dependence for the stochastic variables. It also assumes a
gaussian distribution of the z-space of the ADS-B data which can lead to mediocre results depending on the given data.

Compared with the presented models, the approach developed in this paper is a deterministic uneven auto-encoder using a single encoder to create a latent representation of the ADS-B data linked to several decoders, each getting different data chosen thanks to a
discriminating feature. This idea was used by~\citet{Yook2020} to separate the sound received by speakers placed differently but to the best of our knowledge, was never used in the anomaly detection field. As a result, the latent space created from the single encoder well represents the ADS-B data while the different specialized decoders well capture the information, addressing the variability of the time series over certain period of time, resulting into better detection.

\section{DAE : Discriminatory auto-encoder}
\label{sec:dae}

\subsection{Problem statement}

The task of detecting abnormal ADS-B messages falls into the category of anomaly detection in multivariate time-series. A time series contains successive observations which are usually collected at equal-spaced time-stamp. A multivariate time series $\mathbf{x}$ of length $\mathbf{N}$ is defined as \(\mathbf{x} = \{\mathbf{x_1},\mathbf{x_2},\dotsc,\mathbf{x_N}\}\), 
where an observation \(\mathbf{x_t}\in \mathbf{x}\)  is an \(M\)-dimensional vector at time \(t (t \leq N),\ i.e.\ \mathbf{x_t} = [x^1_t,x^2_t,\dotsc,x^M_t]\) such that \(\mathbf{x} \in R^{M \times N}\). 
The dimension \(M\) represents the number of features in an observation  \(\mathbf{x_t}\). In the domain of anomaly detection in time series, the goal is to find out if an observation \(\mathbf{x_t}\) is anomalous or not. 
However, time windowsare usually preferred to single observations in order to get a better understanding of the evolution of the data over time. 
A Time window \(\mathbf{W_{t-T:t}}\in R^{M \times (T+1)}\) is a set of \(T+1\) observations \(\{\mathbf{x_{t-T}},\mathbf{x_{t-T+1}},\dotsc,\mathbf{x_t}\}\) 
from time \(t-T\) to \(t\). The goal is then to determine if a particular time window is anomalous or not.

Even though time windows always come from the same time series, some external contextual factors may alter the shape of the time windows over time. For instance, Figure \ref{fig:phase} clearly shows that ADS-B time windows created out of a single flight will have significant differences depending on the phases they are taken from.
Hence, every time window \(\mathbf{W_{t-T:t}}\) is associated with a discrimination feature \(\mathbf{D_{t-T:t}}\) to address these discrepancies. The goal of this feature is to mark differences between time windows whether it is time wise, nature wise etc. This can be seen as a static feature that is used by the model in the likes of~\citet{Miebs2020} but which is not a part of the training per se. For instance the flight phases from which ADS-B time windows are taken are used as discrimination feature in this paper.

\begin{center}
\includegraphics[width=0.9\columnwidth]{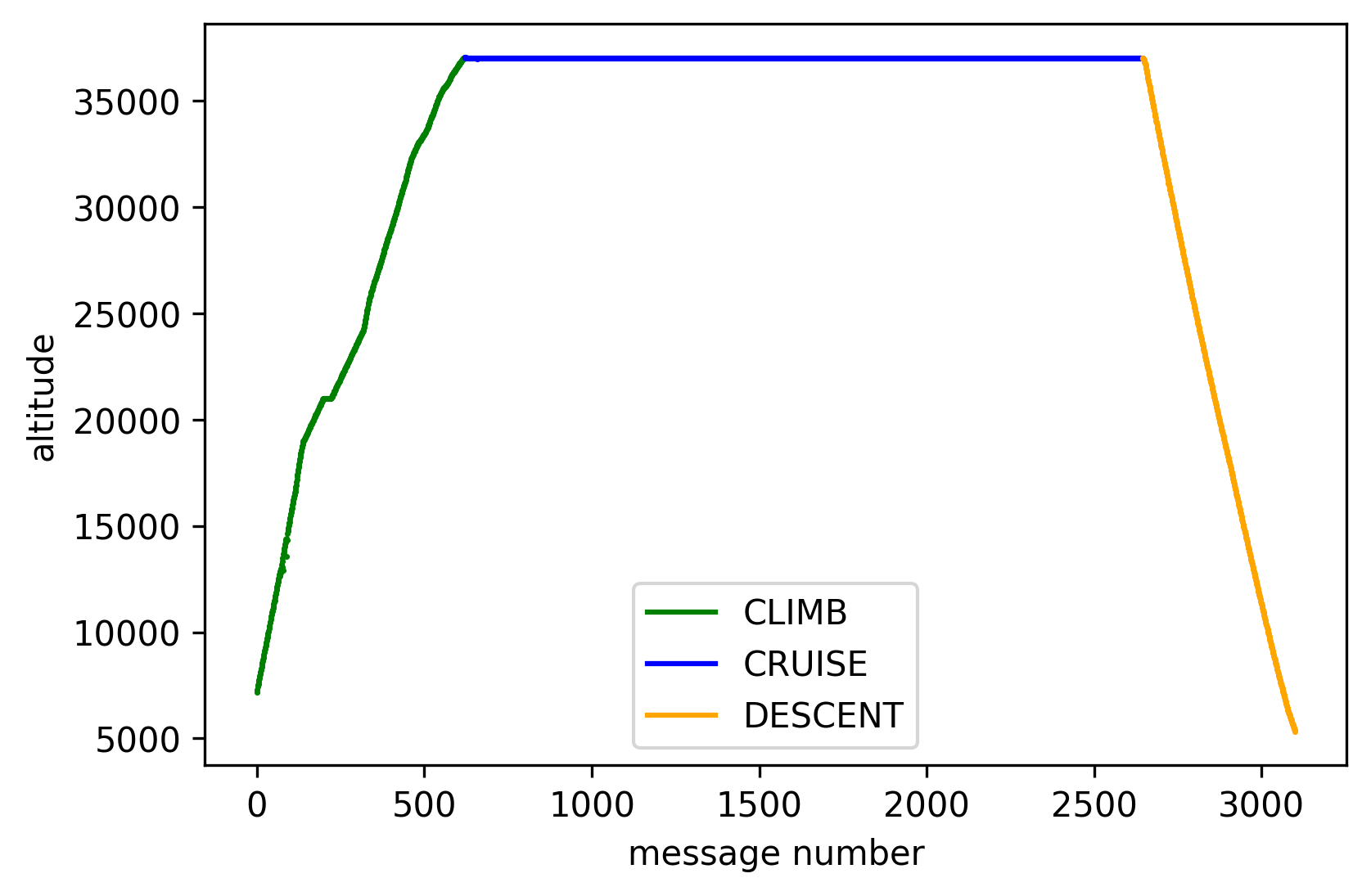}
\captionsetup{justification=centering}
\captionof{figure}{The 3 different phases of a flight used as discriminating feature}
\label{fig:phase}
\end{center}

\subsection{Model architecture}

\begin{figure*}[!ht]
\center
\includegraphics[width=\textwidth]{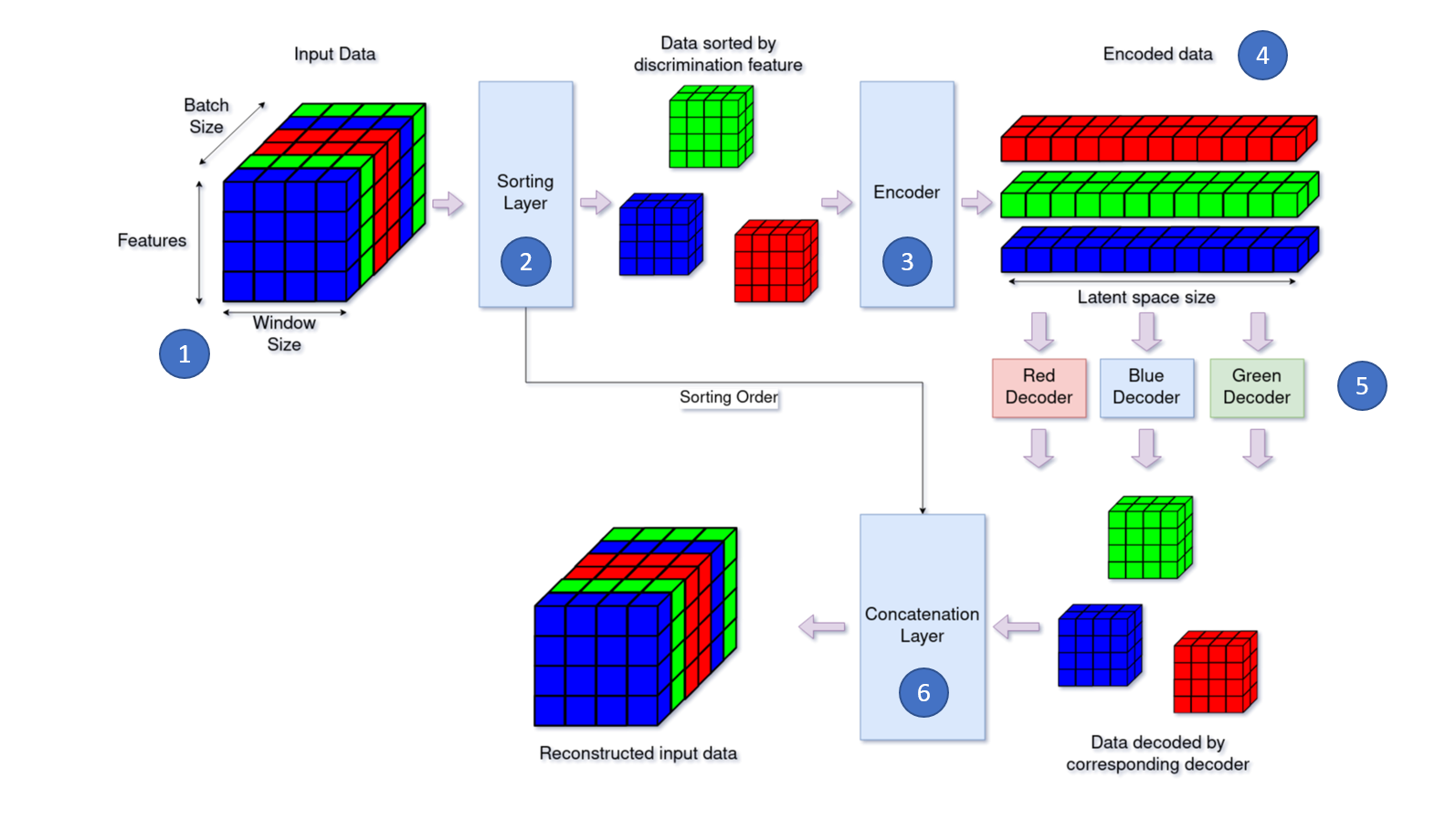}
\caption{Architecture of the DAE model}
\label{fig:model}
\end{figure*}



The basis of the DAE model itself uses the architecture of a classic auto-encoder model~\citep{LIOU201484} made of an encoder and a decoder. The main difference is its unbalanced numbers of encoder and decoder depending on the discriminating feature presented previously. This section explains the different parts of the model that can be found in the Figure \ref{fig:model}.

\vspace*{.15cm} \noindent\circled{1}~\textbf{Input Data:} The input data are constituted of 2 parts. On one hand, the multivariate time-windows are the actual data entering the model. These are 3-dimensional arrays, shaped to be used in RNN layers. On the other hand, the discriminating data is a one-dimensional array which associate each time-window to a discriminant feature.

\vspace*{.15cm} \noindent\circled{2}~\textbf{Sorting Layer:} The sorting layer separates the batches of windows into mini-batches according to the discrimination feature. Each mini-batch is then encoded seamlessly. The sorting layer sends the order of the different windows to reconstruct the batch as is to the concatenation layer. The number of mini-batch depends on the number of values the discriminant can take. For instance, the discriminating feature used in the evaluation of this paper is the phase of the flight which is set to 3 (ascending, cruising and descending).

\vspace*{.15cm} \noindent\circled{3}~\textbf{Encoder:} One can use recurrent neural networks (RNN) to address the time aspect of the data. The main problem of classic RNNs is their struggle to learn the long-term dependencies in a sequence because of the gradient vanishing during learning. ADS-B time windows can be up to 60 seconds long and the model must remember what the state of the aircraft was in this time span. Alternatives to RNNs are LSTMs and more recently the GRU, which do not suffer from the vanishing gradient problem thanks to a system of gating units. In most cases these variants perform equally, and while GRU can have less parameters on smaller dataset, LSTMs having a separate update gate and forget gate can be more effective on longer sequences than the GRU.
To add up additional context to the latent representation of the ADS-B time windows, a bidirectional mechanism is added in the encoder layer in order to use both close past and future to encode the data. 

\vspace*{.15cm} \noindent\circled{4}~\textbf{Latent Space:} The latent representation captures the normal patterns of the ADS-B multivariate time series, considering their time dependence thanks to the LSTM used in the encoding network. The dimension of this vector is important in the DAE as a small value would likely underfit the input time series while a larger one would increase drastically the training time of the model. The dimension is usually smaller than the original number of features found in the input data but in the case of the DAE, the time dependency itself need to be taken into account, explaining a larger size dimension in the latent space than the input. Precise dimension used during experimentation is showcased in the Section \ref{sec:eval}

\vspace*{.15cm} \noindent\circled{5}~\textbf{Discriminated Decoders:} The decoding part of the DAE is mainly what makes it different from a regular auto-encoder. While the encoder encodes all the mini-batches yield by the sorting layer seamlessly at the same time, the decoding part is carried out by several decoders, one per mini-batch. This leads to specialized decoders depending on the data they received during their training. Compared to the encoder, the decoders are composed only of a single LSTM layer and not a Bi-LSTM as it was not deemed important for the decoding since the information was already included in the latent representation. As a result, the DAE is asymmetrical in two ways: the numbers of encoders and decoders are different and the encoder is overall bulkier than the decoders. This is justified by the importance of the quality of the encoder knowing it is alone to complete its task.

\vspace*{.15cm} \noindent\circled{6}~\textbf{Concatenation Layer:} this layer uses the order kept in memory by the sorting layer to reconstruct the data w.r.t. its original order. It is solely used for the training due to the use of different thresholds for each decoder for the anomaly detection, making this concatenation layer obsolete once training is over.

\subsection{The thresholds calculation and the anomaly detection}

Once the data is reconstructed by the decoders, the input and the output are compared to calculate a similarity score.
The higher this score is, the better the model managed
to recreate the input time series. Here, each time window \(\mathbf{W_{t-T:t}}\) gets its own reconstruction score calculated by, for instance, a mean squared error. After training, instead of using directly the reconstruction score, an anomaly score is defined as:

\begin{equation} \label{eq:MSE}
Anomaly(\mathbf{W_{t-T:t}}) = \frac{1}{n}\sum_{i=0}^{n}1 - ({\mathbf{x}_i-\hat{\mathbf{x}}_i})^2
\end{equation}

This score is compared against a threshold to determine if the window associated to it contains an anomaly or not.
As discussed in the previous section, the model has different decoders trained on different data. As a result, the
loss of each decoder is going to be different,
leading to anomaly scores not being equivalent across the different
decoders. In this context, having the same anomaly threshold for all the decoders would be counterproductive and lead
to mediocre metrics. 

Calculating the threshold can be done in several ways. \citet{Luo2021} uses support vector data description (SVDD) to
determine automatically the best threshold to use. SVDD is an unsupervised model that creates boundaries around the
training dataset that is then used on testing data to determine whether they are out of bounds (i.e. anomalous). 
It is usually trained using a mix of positive and negative to make it more robust to outliers contained in the
training set. Unfortunately, in the case of ADS-B, not only real life anomalous examples are scarse, but the data also
tend to contain outliers due to already discussed problems which make an SVDD hard to use successfully on real
not-over-processed data.

For simplicity and efficiency, the 3-sigma rule is used to calculate the threshold for the DAE. Considering each decoder has its own output distribution, the calculation for the threshold is done on the training data for each one of them. The threshold \(\tau\) is defined as \(\tau = \mu + 3\sigma\) where \(\mu\) is the mean of the anomaly score distribution of one decoder and  \(\sigma\) is its standard variation. This results in having a threshold value being different depending on the decoder the data went in. 

The distribution of the anomaly scores, which roughly following a normal distribution (see Figure \ref{fig:hist}), assures the 3-sigma rule to yield a low false positive rate while being sensitive enough to flag anomalies.

\begin{center}
\includegraphics[width=0.9\columnwidth]{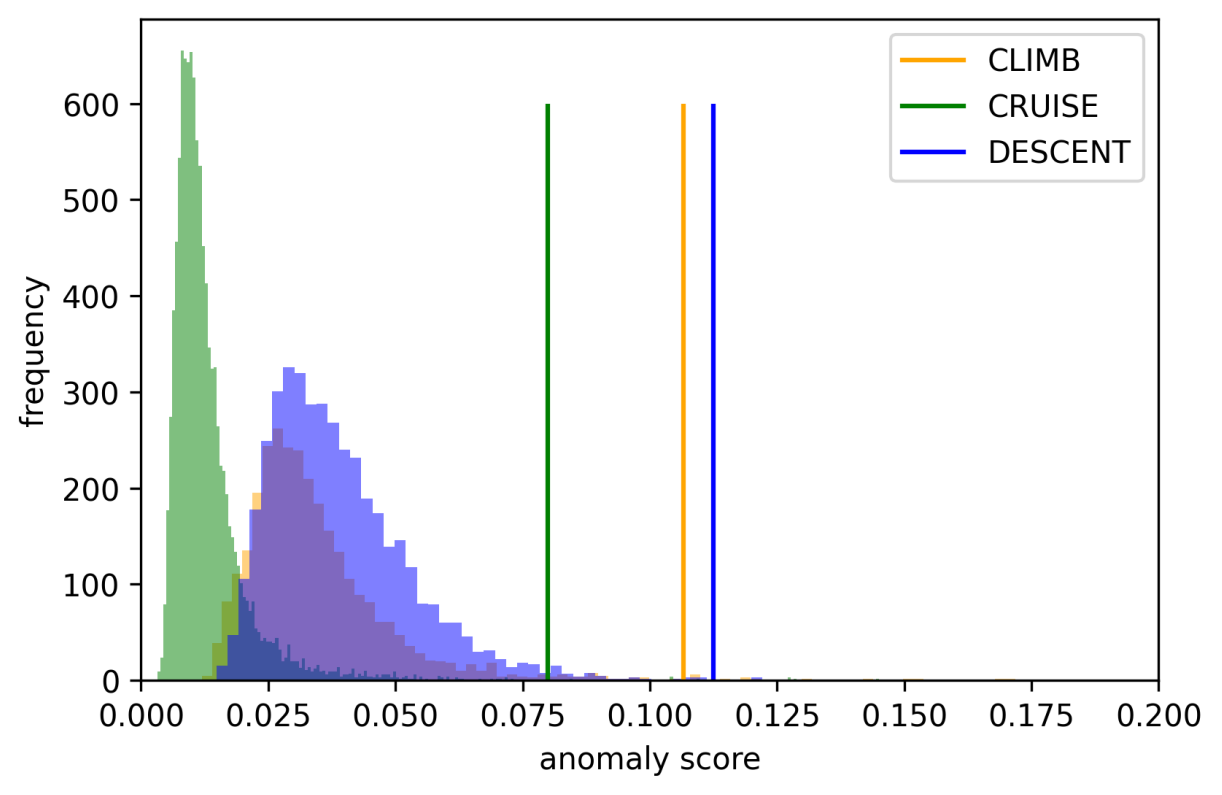}
\captionof{figure}{The anomaly score's distributions of the training data for each phase along with its calculated thresholds}
\label{fig:hist}
\end{center}

\section{Dataset creation and pre-processing}
\label{sec:data}


Apart from the DAE, another contribution of this paper is the availability of a dataset to train and evaluate multivariate anomaly detection models. This section presents the different tools used for the creation of the dataset as well as the pre-processing to obtain the final data.

\begin{center}
\includegraphics[width=\columnwidth]{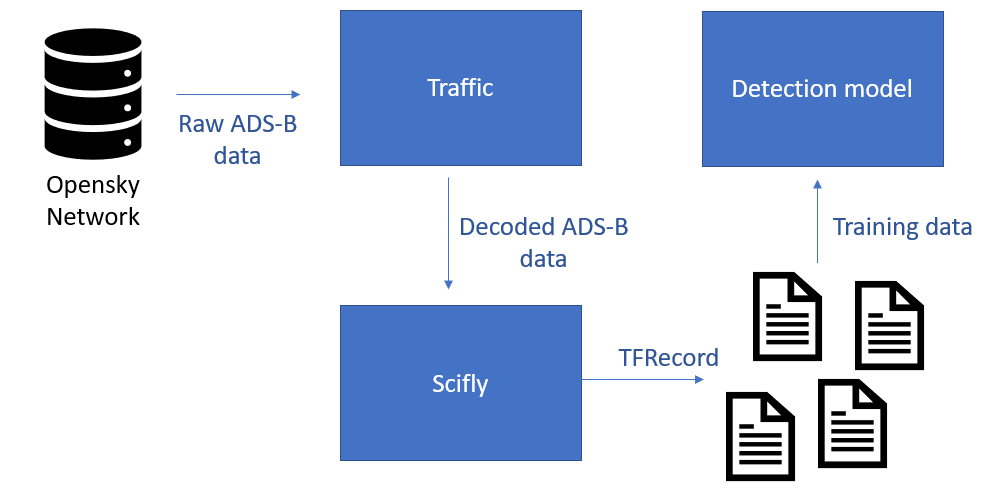}
\captionsetup{justification=centering,margin=0.5cm}
\captionof{figure}{The data architecture used for gathering and processing ADS-B messages into training data.}
\label{fig:data_archi}
\end{center}

\subsection{Global architecture}

In Figure \ref{fig:data_archi} is described the overall architecture used to train the DAE. After retrieving data from the Opensky Network, the data pre-processing cleans the data, getting rid of
aberrations caused by decoding errors or sensor inaccuracy and then creates the different features needed for training.
Through serialization, the processed data are then sent to the Model training block which creates the windows and
standardizes the data inside the Tensorflow's data.io framework to stream the data directly during the training using
tfrecord format. The model learns the normal patterns of flights through regular historical data and outputs an anomaly
score for each time windows. These normal anomaly scores are then used to determine a threshold which separate
normal data from anomalous ones using the 3-sigma rule. The model can then be used on unseen data to determine their nature by calculating its score. If the score is under the threshold, then the time window is considered normal, else,
it is considered abnormal. 

\subsection{Data acquisition}


\vspace*{.15cm} \noindent\circled{1}~\textbf{Opensky Network} is an online flight tracking network which 
provide access to data collected by cooperative ground stations. The large-scale dataset of this evaluation is
historical data extracted from their historical database. 

\vspace*{.15cm} \noindent\circled{2}~\textbf{Traffic}~\citep{Olive2019a} is an 
open-source Python-based tool allowing users to query the Opensky Network historical 
database. It simplifies the data gathering by aggregating the different types 
of ADS-B messages (position, velocity and identification) as well as making the 
data cleaning less cumbersome. The Opensky Network data comes in two main different forms available to the 
user : one in a form of already processed and cleaned data called state-vectors 
and the other in raw messages in BEAST format. 
While the former would be a time-saver, it would not yield full control over the 
data preparation. On the other hand, getting the raw transmission not only 
gives more freedom to the user but also allows experiments using directly 
an ADS-B feed delivered by a private antenna. With recent iterations of Traffic, the processing of raw 
data has become much easier. With the implementation of a clustering algorithm 
used by~\citet{sun2017flight}, the raw data which consist of series of 
messages, once decoded, can be separated into well defined flights.

As mentioned earlier, the relevant ADS-B messages are sent from position, velocity and identification messages. These
messages are, according to the ADS-B specification, sent by the transponder every half-second for the position and
velocity messages and once every five seconds for the identification ones. This very short timeframe between receptions
leads to a consequent amount of data with a non-negligible redundancy. Traffic allows for the downsampling and the
concatenation of the different messages. The original timeseries are then transformed from several messages per seconds
down to one every two seconds. This has the clear advantage of reducing the size of the dataset without losing
meaningful data due to the high redundancy implied by the high-rated emissions. Another valuable gain from this
reduction of messages is the amount of information contained into a time window. 
Indeed, without the re-sampling, a time window of 30 messages would be equivalent to around 13 seconds of recording.
Changing the original rate of messages to 2 seconds bring the 30 messages window to a full minute of recording.
This strongly impacts both the training time of the model as well as its accuracy as it improves the time dependencies developed by the RNN layers.  

\vspace*{.15cm} \noindent\circled{3}~\textbf{Scifly} is a toolbox additional
to Traffic\footnote{https://github.com/Wirden/scifly} developed in the context of the current work. Despite using data from a well-covered area like Europe,
errors in decoding the data or approximation from sensors  still happen which often result in big leaps of the aircraft
during a flight. These corrupted data are undesired in training dataset and would result
in lower quality models. 
To filter out these blatant outliers, we check the distance in kilometers between close neighbour messages (consecutive ones) and separated messages. 


The other use of Scifly is to export the ADS-B data in TFRecord. TFRecord format uses protocol buffers
\footnote{https://developers.google.com/protocol-buffers/} to serialize data making it available to a large share of
machine learning algorithms. It also have the advantage to allow to shard the data in multiple files to parallelize the
input data for optimize training. Lastly, Scifly allows the exporting and importing of data to and from FDI-T, our
anomaly creation platform.


\subsection{FDI-T}

\emph{FDI-T} is a testing framework that we developed~\citep{Vernotte2021} jointly with
  Smartesting (\url{https://www.smartesting.com}) and Kereval
  (\url{https://www.kereval.com/}). It
allows ATC experts to design FDIA scenarios to alter (i.e. create,
modify and delete) recorded legitimate ADS-B surveillance messages.
The altered recordings can then be played back (w.r.t. time
requirements) onto real surveillance systems or can be exported e.g. to train and/or
validate Machine Learning models. The goal is to simulate an attacker
tampering with the surveillance communication flow.

The types of alteration to apply are specified through the 
definition of alteration scenarios, of which the design is textual-based via a Domain Specific Language (DSL). Once designed, the scenarios are automatically applied on source recordings of air traffic surveillance communications, thanks to a dedicated alteration engine~\citep{Cretin2020}.
Alteration scenarios have various parameters, such as a time window, list of targeted aircraft, triggering conditions, and others parameters related to the alteration's type. All parameters are recording agnostic, meaning that scenarios can be applied to multiple recordings regardless of their nature. All these features truly make the creation of ML dataset a fast albeit precise procedure.

Concretely, FDI-T was used in this study to create many of the scenarios that constitute the evaluation dataset.


\begin{table*}[!htb]
\setlength\tabcolsep{0pt} 
\centering
\begin{tabular*}{0.9\textwidth}{@{\extracolsep{\fill}}llcc}
\toprule
Departure airport & Arrival airport & Number of flights & Duration (hours) \\
\midrule
  Athens (LGAV)     & London (EGGW)     & 56    & 3.6 \\
  Berlin (EDDB)     & Kiev (UKBB)       & 33    & 1.6 \\
  Budapest (LHBP)   & Dublin (EIDW)     & 43    & 2.8 \\
  Frankfurt (EDDF)  & Lisbon (LPPT)     & 68    & 2.5 \\
  Hamburg (EDDH)    & Barcelona (LEBL)  & 29    & 2.0 \\
  Kiev (UKBB)       & Paris (LFPG)      & 83    & 3.3 \\
  London (EGGW)     & Milan (LIMC)      & 46    & 1.6 \\
  Madrid (LEMD)     & Moscow (UUEE)     & 59    & 4.2 \\
  Malaga (LEMG)     & Frankfurt (EDDF)  & 81    & 2.9 \\
  Manchester (EGCC) & Istambul (LTFJ)   & 75    & 3.8 \\
  Munich (EDDM)     & Lisbon (LPPT)     & 68    & 3.3 \\
  Paris (LFPG)      & Oslo (ENGM)       & 34    & 1.9 \\
  Stockholm (ESSA) & Barcelona (LEBL)  & 25    & 3.2 \\
  Vienna (LOWW)     & Copenhagen (EKCH) & 83    & 1.3 \\
  Zurich (LSZH)     & London (EGLL)     & 225   & 1.2 \\
\midrule
  Hamburg (EDHI)    & Hawarden (EGNR)   & 45    & 1.5 \\
  London (EGLL)     & Moscow (UUEE)     & 184   & 4.0 \\
\bottomrule
\end{tabular*}
\captionsetup{justification=centering,margin=1cm}
\caption{Flights used for the training of the different models presented. 15 flight routes data taken from September to December 2020 for training and 2 flight routes taken in January 2021 for validation.} 
\label{tab:train}
\end{table*}

\subsection{Training data}

The different flight routes used for the training can be visualized in Table \ref{tab:train}. It compiles together 15 flight routes for a total of 1008 flights. The training dataset is exclusively focused on internal European flights. This choice is motivated, mainly, by the excellent land coverage of the Opensky Network in this area. This ensures good quality data without major discrepancies due to low quality ground station or non-covered area. The dataset is composed of both long and short flights, as well as flights traveling in different directions to ensure data diversity.
From the data gathered through the described architecture, only some features of ADS-B messages are kept and fed to the model: 

\begin{itemize}
\item \textbf{Altitude} in feet given by the airborne position messages. 

\item \textbf{Consecutive Delta} in kilometers. This is the Vincenty distance 
between two consecutive messages calculated from the latitudes and longitudes. 
This distance is bound to change from 2 main factors. 
The first one is the change of speed of the aircraft and the second 
one is the absence of messages picked up by the OpenSky Network. The third reason
would be errors in decoding or from sensor malfunctions but most are filtered out
from the data cleaning processing explained above.


\item \textbf{Tracking Delta}. Difference between the tracking received through ADS-B
and the \emph{ideal} tracking calculated from the position of the aircraft and the
position of the arrival airport.
\item \textbf{Vertical Rate} in feet/mn. Represents the aircraft's vertical 
speed – the positive or negative rate of altitude change with respect to time.
\item \textbf{Groundspeed} in knots. Represents the speed over ground.

\item \textbf{Phases}. Categorical feature used as the discriminating feature to choose the decoder. The fuzzy logic developed by~\citet{sun2017flight} is used to automatically determine the phase a window is originated from. In addition, a rule has been added forcing the cruising phase when the flight is over 300 kilometers away from the departure or arrival airport. This helps when the fuzzy logic labels a crash as a simple descending maneuver. Figure \ref{fig:phase} shows the different phases of a flight automatically determined by the Scifly algorithm.

\end{itemize}

It is worth noting that some base features of ADS-B like the tracking, the latitude
and the longitude are not directly used in the dataset. Concerning the tracking, the
feature being a cyclic feature in degree, experiments were made using the sine and
cosine component to avoid the discontinuity implied by having a heading varying between
0 and 360 when 0 and 360 being de facto the same angle. Unfortunately, having two features
for the heading instead of one doubled its impact on the model and created some unbalance
hence the choice for the heading delta feature presented earlier.

In a similar fashion, the latitude and longitude also being cyclical data were turned into
the consecutive delta feature. Another reason for this change is the will to make the model
area-agnostic which would have been impossible with the coordinates as is as features. This will
improve the accuracy of the models on data they have not seen during their training, as shown by
\citet{Fried2021}.



\subsection{Evaluation Data}

\begin{figure*}[!ht]
\center
\includegraphics[width=0.6\textwidth]{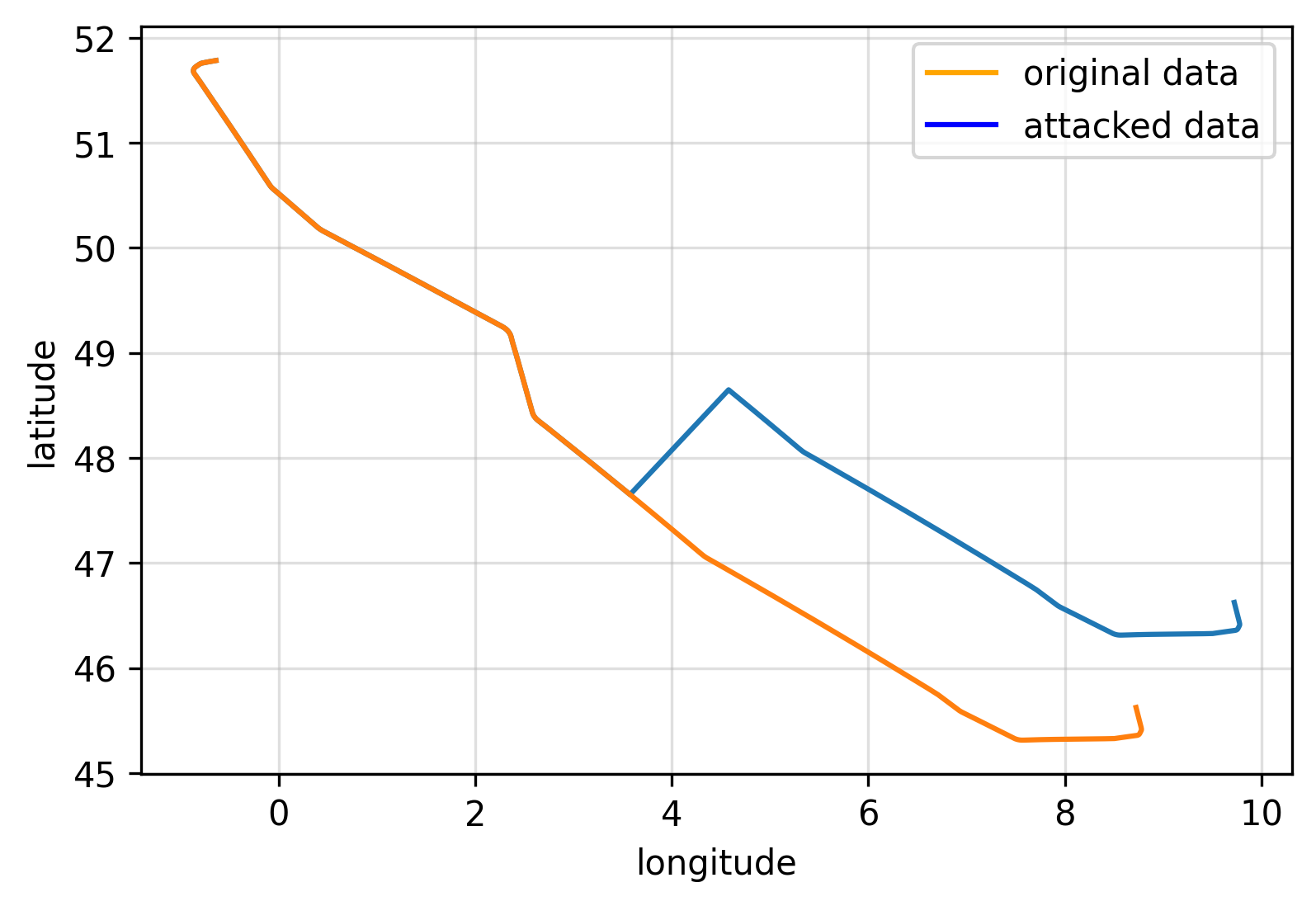}
\captionsetup{justification=centering,margin=2cm}
\caption{Constant position offset attack}
\label{fig:offset}
\end{figure*}

\begin{figure*}[!ht]
\center
\includegraphics[width=0.6\textwidth]{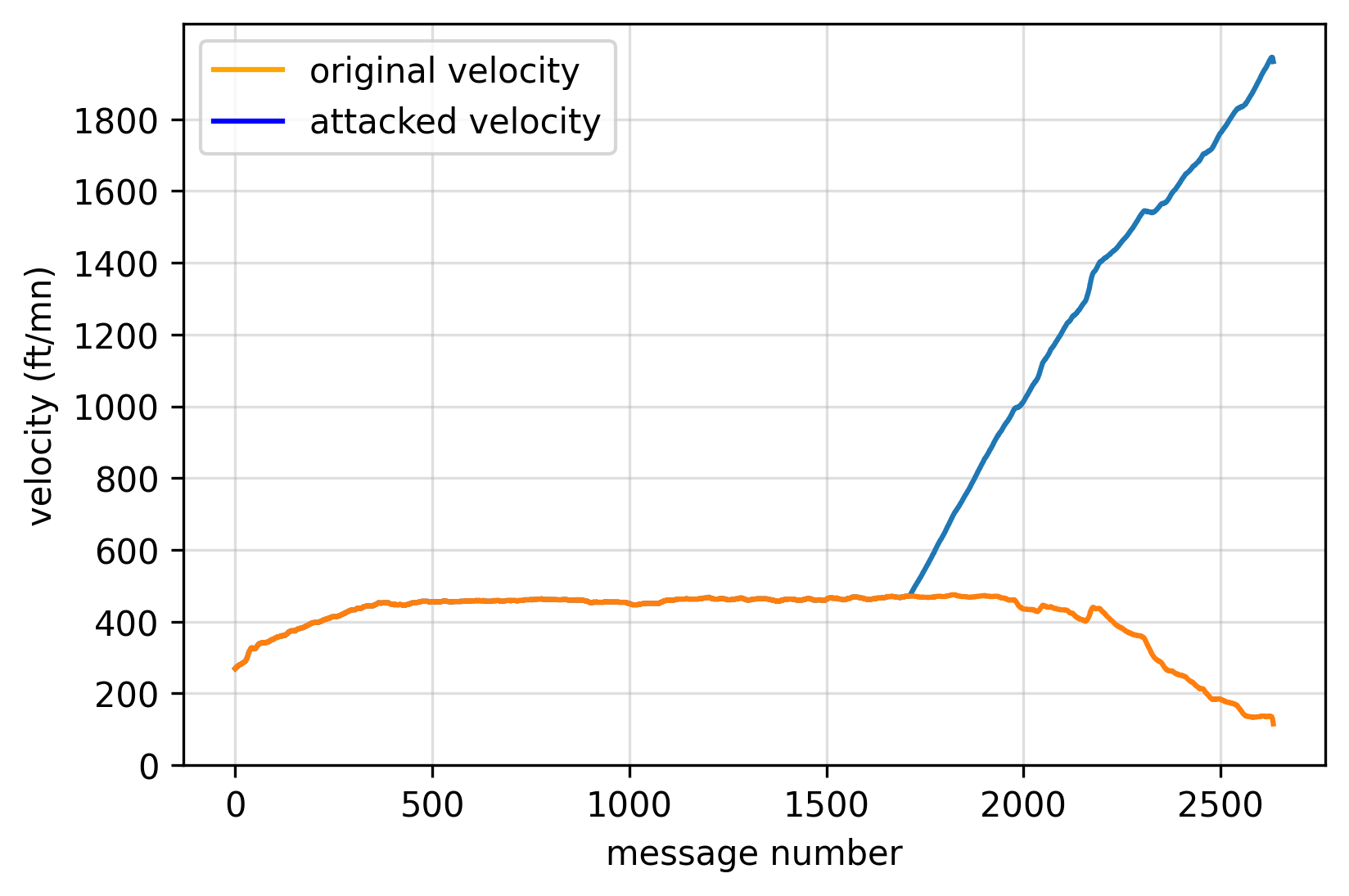}
\captionsetup{justification=centering,margin=2cm}
\caption{Velocity drift attack}
\label{fig:drift}
\end{figure*}

\begin{figure*}[!ht]
\center
\includegraphics[width=\textwidth]{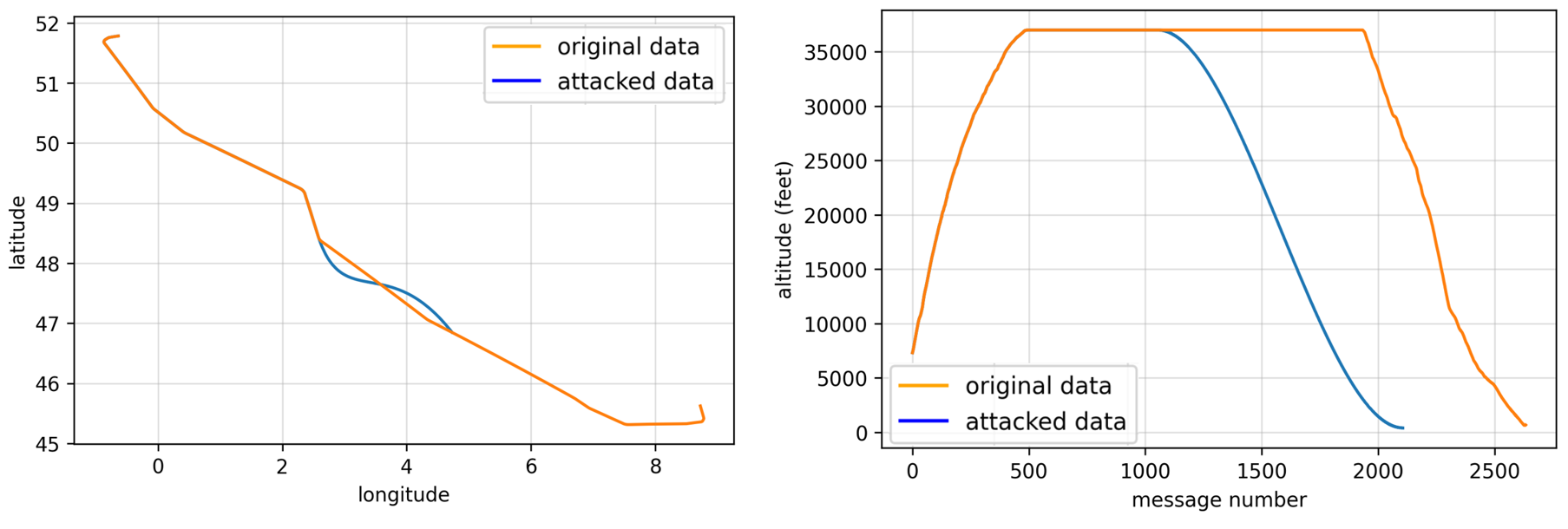}
\captionsetup{justification=centering,margin=2cm}
\caption{Crash attack. Latitude / longitude on the left and the altitude on the right. Other features like vertical speed or track are also modified \emph{realistically}.}
\label{fig:crash}
\end{figure*}

The evaluation dataset is composed of different scenarios to test different situations an anomaly detection model could be used in. It is a mix of regular data taken from real life and anomalies created or not using the FDI-T framework. All the anomalies, except the one already found in real data, are applied on all the flights found in the training dataset found in Table \ref{tab:train_data} from January 2021 instead of the last 3 months of 2020:


\textbf{World Data (WORLD)} -- As the training dataset is exclusively composed of data from European flights, including
regular data from other parts of the world in the testing dataset allows for checking the genericity of the approach. It
includes flights from the european airspace, american airspace -- e.g. Dallas to Louisville -- or australian airspace
-- e.g. Camberra to Perth --.

\textbf{Gradual Drift (DRIFT)} -- Anomalies that consist of simulating an
altitude drift or a velocity drift. The altitude or velocity messages on the attacked time window are
all raised/lower by an increasing/decreasing multiple of $n$ feet. So, if
the first message is lowered by 25 feet, the second will be
lowered by 50, etc. The Figure \ref{fig:drift} shows a velocity drift used during the evaluation

\textbf{Made-up Crashes (CRASH)} -- Using FDI-T, life-like crashes scenarios can be created combining an altitude drift, a negative vertical rate, and a reduction of groundspeed -- not to be mistaken with airspeed --. The signal is then stopped once the aircraft lands. Figure \ref{fig:crash} shows some of the features modified during a CRASH attack.

\textbf{Ryanair Hijack (HJK)} -- Constituted of the Ryanair flight 4978 from Athens to Vilnius which was forcibly diverted to Minsk after entering the Belarus airspace on the 23rd of May 2021\footnote{https://www.flightradar24.com/blog/ryanair-flight-4978-to-vilnius-forcibly-diverted-to-minsk/}. It is to be noted that the emergency was turned on by the crew 2 minutes after the flight started to change its course. For the evaluation, the labels have been set to 1 from the beginning of the emergency till the landing.

\textbf{Constant position offset (OFFSET)} -- This ano\hyp{}maly, when toggled takes the real data of a flight and adds an offset of 1 in both the latitude and the longitude (see Figure \ref{fig:offset}). This offset represents a distance of around 132 kilometers between the original and the anomalous trajectory.

\section{Experimental Evaluation}
\label{sec:eval}

This section presents the experimental evaluation of the model presented above. It also display the differences in performance between the DAE and other anomaly detection methods.

\subsection{Model training specifications}

The DAE model was trained with the training dataset above-mentioned which represent 336 Mo of data separated
into 15 tfrecord files. Tensorflow interleaves the data contained in these files to feed it to the model
during the training avoiding risks of memory overflows. The training was made on the Mésocentre de Calcul
de Franche-Comté using a Tesla V100 performing at 7.8 TeraFLOPS. The training on average
was taking around 26 minutes per epoch. In Figure \ref{fig:train_loss} a difference in loss between the training set and the
validation set can be observed. It is due to a few outliers in the training set which make the average training 
loss way higher than its validation counterpart.

\begin{center}
\includegraphics[width=0.9\columnwidth]{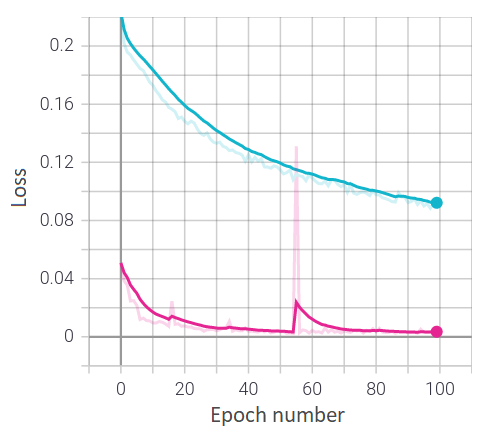}
\captionsetup{justification=centering}
\captionof{figure}{The training loss per epoch. In blue is the training loss, in pink the validation loss}
\label{fig:train_loss}
\end{center}

To train the DAE of which the results can be found in the following section, windows of 30 messages have been used. The batch-size is 256. The number of units in the encoder's BiLSTM is 32 which is then flattened to feed a Dense layer reducing the dimension to 10, the chosen latent space size. The different decoders each embed a single LSTM layer with 32 units.
For the other models found in the evaluation, the features, the hyper-parameters and the threshold selection method found in their respective papers were used if provided.  

In order to properly evaluate and compare the different models performance, accuracy (ACC), Recall (R), False Positive Rate (FPR) and F1-score (F1) are used: 

$\begin{cases}
Acc =\frac{\text{TP} + \text{TN}}{\text{TP} + \text{TN} + \text{FP} + \text{FN}} \\
R =\frac{\text{TP}}{\text{TP} + \text{FN}}\\
FPR =\frac{\text{FP}}{\text{FP}  + \text{TN}}\\
F_{1}=\frac{2\text{TP}}{2\text{TP} + \text{FP} + \text{FN}}
\end{cases}$

\noindent where TP, FP, FN and TN being the values found in a regular 2x2 contingency table.

All the results and implementation of this paper are accessible on the Scifly~\footnote{https://github.com/Wirden/scifly} Github repository. The full dataset used for the different models can also be found at this address. This is made as an effort to improve the replicability of the presented evaluation as well as proposing a non-exhaustive, upgradable baseline dataset for future models in the growing field of anomaly detection in ADS-B data.

\begin{table*}[!htb]
\setlength\tabcolsep{0pt} 
\centering
\begin{tabular*}{0.9\textwidth}{@{\extracolsep{\fill}}llcccr}
\toprule
    Evaluation Dataset      & Evaluation metric & LSTM-AE & IForest & VAE-SVDD & DAE \\
\midrule
  World Data                & Accuracy      & 0.994 & 0.687    & 0.899 & 0.989 \\
                            & Recall        & NaN   & NaN   & NaN   & NaN \\
                            & FPR           & 0.006 & 0.313    & 0.101 & 0.011 \\
                            & F1 score      & 0   & 0    & 0   & 0 \\
  Ryanair Hijack            & Accuracy      & 0.946 & 0.890    & 0.722 & 0.847 \\
                            & Recall        & 0.637 & 1    & 0.227 & 0.301 \\
                            & FPR           & 0.001 & 0.129    & 0.231 & 0.017 \\
                            & F1 score      & \textbf{0.778} & 0.729    & 0.152 & \textbf{0.439} \\
  Velocity drift            & Accuracy      & 0.933 & 0.944    & 0.949 & 0.961 \\
                            & Recall        & 0.809 & 0.957    & 0.930 & 0.912 \\
                            & FPR           & 0.001 & 0.063    & 0.043 & 0.012 \\
                            & F1 score      & 0.886 & \textbf{0.937}    & 0.926 & \textbf{0.939} \\
  Constant position offset  & Accuracy      & 0.519 & 0.709    & 0.541 & 0.526 \\
                            & Recall        & 0.033 & 0.491    & 0.077 & 0.053 \\
                            & FPR           & 0.001 & 0.073    & 0.046 & 0.004 \\
                            & F1 score      & 0.060 & \textbf{0.598}    & 0.107 & \textbf{0.097} \\
  Made-up Crash             & Accuracy      & 0.506 & 0.919    & 0.710 & 0.962 \\
                            & Recall        & 0.003 & 0.922    & 0.426 & 0.929 \\
                            & FPR           & 0.001 & 0.084    & 0.037 & 0.004 \\
                            & F1 score      & 0.005 & \textbf{0.925}    & 0.573 & \textbf{0.955} \\
\midrule
  Total                     & Accuracy      & 0.780 & 0.830    & 0.764 & 0.857 \\
                            & Recall        & 0.371 & 0.843    & 0.415 & 0.549 \\
                            & FPR           & 0.002 & 0.132    & 0.092 & 0.010 \\
                            & F1 score      & 0.544 & \textbf{0.797}    & 0.440 & \textbf{0.738} \\                         
\bottomrule
\end{tabular*}
\caption{Comparison of the different models evaluated} \label{tab:eval}
\end{table*}        

\subsection{Compared Results against other ML methods}

\begin{figure*}[!ht]
\center
\includegraphics[width=.9\textwidth]{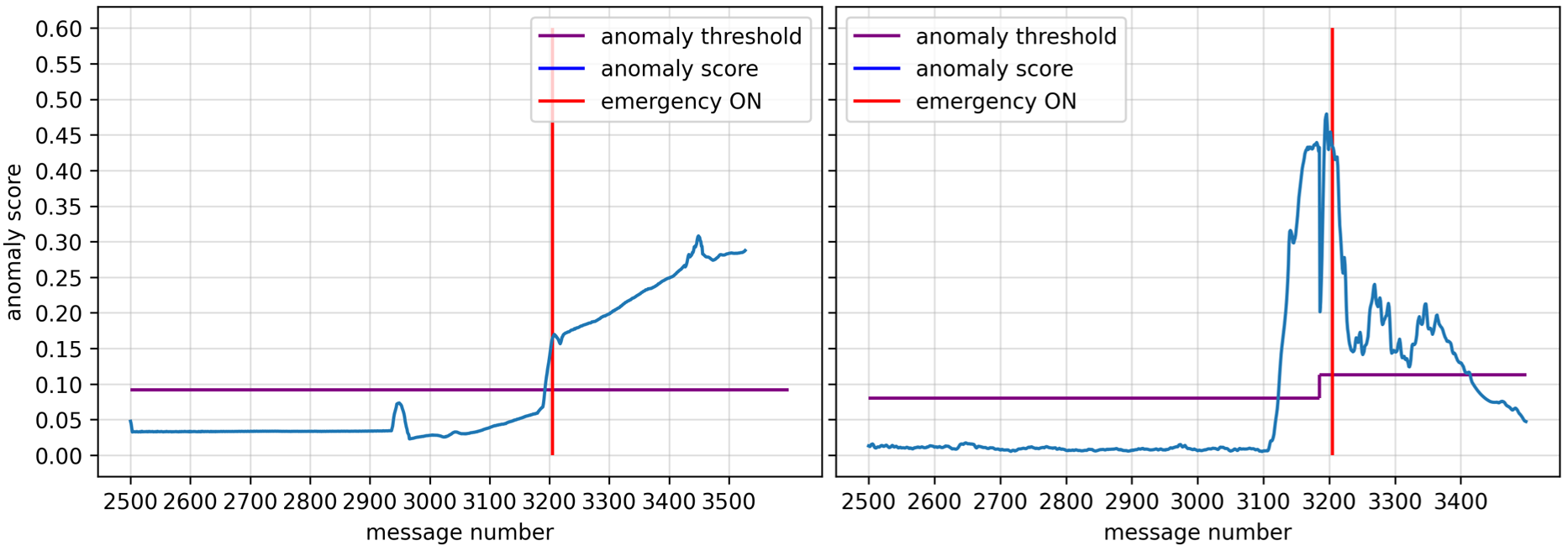}
\captionsetup{justification=centering,margin=2cm}
\caption{Anomaly score for the LSTM-AE on the left and the DAE on the right for the Ryanair hijack flight.}
\label{fig:AE_DAE_hijack}
\end{figure*}

To show the overall performance of the DAE, it is compared with 3 other unsupervised approaches for anomaly detection in ADS-B time-series : a regular Isolation Forest~\citep{Liu2008} model, an LSTM-auto-encoder~\citep{Habler2018}, and a VAE-SVDD~\citep{Luo2021}. Table~\ref{tab:eval} shows the accuracy, the recall, the FPR and the F1 score on the different dataset for each model. For the VAE-SVDD, the method to choose the anomaly thresholds is already given in the paper and the F1 score is calculated accordingly. For the other models, the 3-sigma ruled is applied on the training data to choose the threshold meaning that approximately 99.7\% of the training data anomaly score are under this value. 
Overall, these experimentation results demonstrate the superiority of the DAE compared with the state-of-the-art approaches in ADS-B anomaly detection. Indeed the F1 score on the Total Dataset is more than 20\% over the second best performing model (not considering the IForest for the reasons explained later). It is to be noted that the WORLD dataset does not have any true positives nor false negatives which automatically set the Recall to Nan (division by zero) and the F1 to 0. Next, we analyze the performance of the different methods in detail. \\
\indent \emph{LSTM-AE} is a sequence to sequence model based on a encoder-decoder reconstruction used by~\citet{Habler2018} for anomaly detection in ADS-B time-series. This simple deterministic model well manages to capture the ADS-B normal behaviour in its latent space showing very low FPR  using a 3-sigma threshold as well as decent results on the Velocity Drift dataset. Its very low F1 score on the Made-up Crash dataset can be explained by the data resembling a regular descent trajectory which leads the decoder to reconstruct the data as is. Lowering the threshold to a 2-sigma could help raising the F1 score but would result to a FPR being way too high for anomaly detection. From the metrics alone, the DAE seems to be under-performing compared to the LSTM-AE for the hijack anomaly. This can be explained by looking at Figure \ref{fig:AE_DAE_hijack} which compares the anomaly score over the message windows for both models. One can observe that for the DAE, the anomaly is set off before the actual emergency due to its delay with the diversion of the flight. It explains the FPR being way higher than the other models and displays the reactivity of the DAE in such circumstances. On the other hand, the low recall is due to the score going back to a \emph{normal} value after some time which means the DAE does not label the end of the flight as abnormal from its ADS-B data. \\

\begin{figure*}[!ht]
\center
\includegraphics[width=\textwidth]{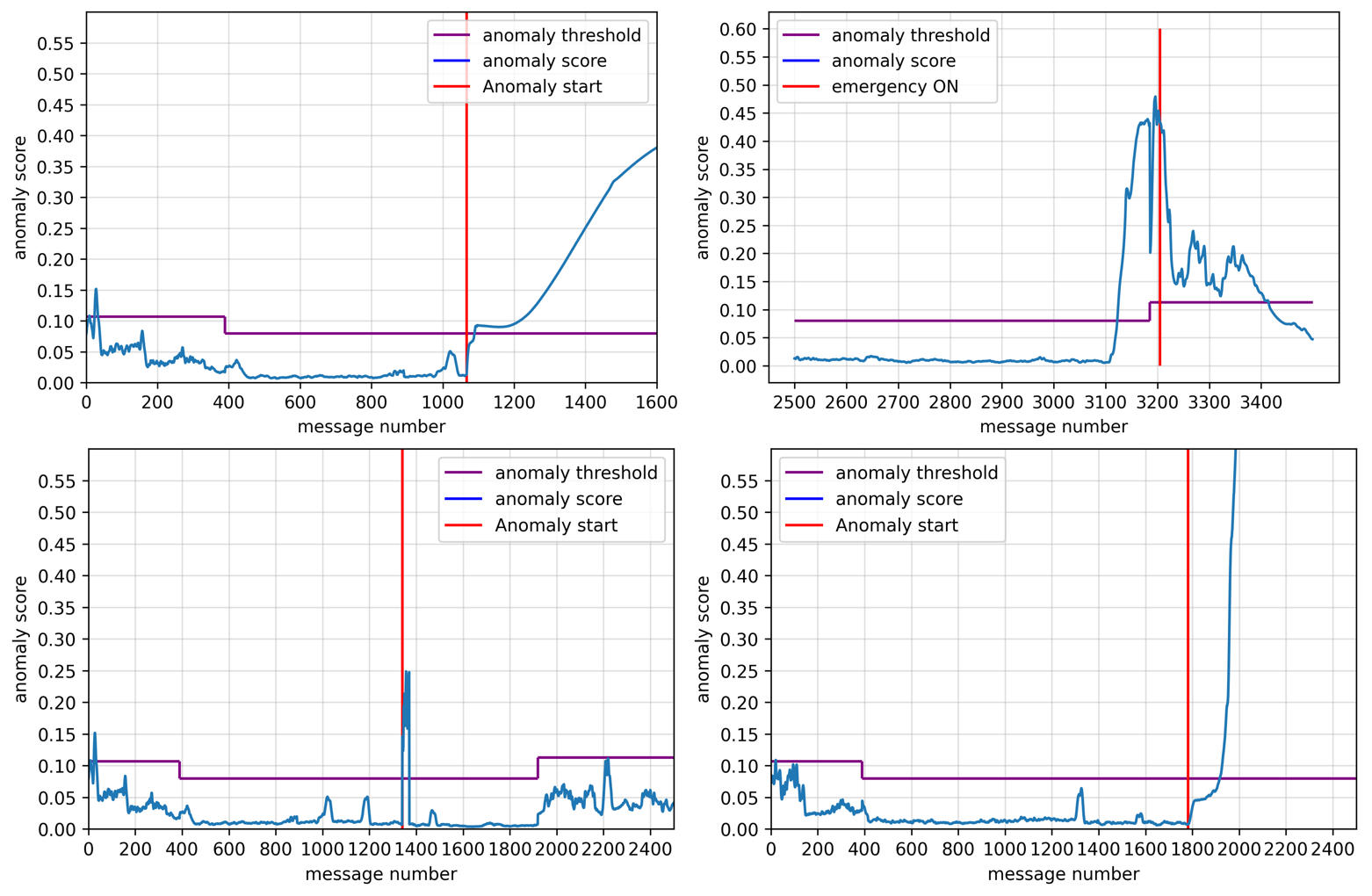}
\captionsetup{justification=centering,margin=1cm}
\caption{DAE anomaly scores for a flight taken randomly from the different evaluation dataset. The top-left figure is from the CRASH dataset, the top-right is from the Ryanair hijack, the bottom-left is from the constant position offset and the bottom-right is from the velocity drift dataset}
\label{fig:dae_results}
\end{figure*}

\indent \emph{VAE-SVDD} is a variationnal auto-encoder (VAE) coupled with a support vector data description model (SVDD) to automatically determine its threshold. A VAE is a deep Bayesian model which represents an input \(\mathbf{x_t}\) to a latent representation \(\mathbf{z_t}\) with a reduced dimension, and then reconstructs \(\mathbf{x_t}\) by \(\mathbf{z_t}\). The main difference with a regular auto-encoder is that the latent variable \(\mathbf{z_t}\) is sampled from a probability distribution, such as a Gaussian distribution with the mean and the standard variation being outputs of the encoder network. This stochastic approach could explain the better results compared to the regular LSTM-AE on the Made-up Crash and Velocity drift dataset. \citet{Luo2021} combine LSTM and VAE by replacing the feed-forward network in a VAE to GRU but do not include information from \(\mathbf{z_t-1}\) into \(\mathbf{z_t}\) in the likes of~\citet{Su2019}. That might explain the issues the VAE-SVDD has to properly represent the distributions of the input data, leading to high FPR compared to the other methods. All in all, the VAE-SVDD, while performing well on coarse anomalies like the velocity drift and to some extents the Made-up Crash thanks to its stochasticity, fails to reconstruct properly ADS-B data leading to high FPR on new data and mediocre results overall. This could be explained by the limitation of having a Gaussian \emph{qnet} being too simple to properly reconstruct ADS-B information coming from other parts of the world, negating the advantages of having such an architecture. \\

\indent \emph{Isolation Forest} is an anomaly detection algorithm using an ensemble of isolation trees to differentiate normal data from anomalies. It has the advantage of being fast, light-weight and can be quickly implemented. 
The model yields good results when compared to the other models, which is explained by the evaluation dataset being based on the same flights as the training data but one month later. The IForest manages to flag anomalies on flights it has already seen or in the vicinity of these said flights -- for instance, the Ryanair Hijack -- without any trade-off except its FPR. Indeed, the FPR is on average almost ten times higher than the DAE's which makes it hard to use as a reliable anomaly detector. It would even be completely pointless on flights in part of the world it did not see during its training.

\indent \textbf{Summary}. Compared to the LSTM-AE with a single decoder, the DAE, thanks to its specialized decoders, manages to discriminate anomalous situations like crashes from regular descent operations while keeping a very similar low FPR overall. Compared to the VAE-SVDD, the DAE performs better on all dataset except on the constant position offset where all models perform poorly due to the scenario's very nature. Indeed, the small offset added to the latitude and longitude is not enough to trigger alarms leading to extremely low F1 scores. This anomaly can only be detected by the LSTM or GRU based models when the values are changed. Finally, the IForest model, despite being cost-effective and accurate on the few flights it has seen during its training, is not as dependable as the LSTM-AE or the DAE, limiting its usage in real-life applications.

\section{Discussion}
\label{sec:disc}

The DAE model shows good results on the chosen evaluation dataset compared to other ADS-B anomaly detection models. Here are some discussion points and caveats for using and improving the model in future works:

\begin{itemize}

\item The first assumption made for the usability of these models is the authenticity of the data used during the training. If data sources like sensors or the Opensky-Network were to be attacked, the models trained from these corrupted sources would not be able to detect ADS-B anomalies properly. 

\item Flight trajectories, while being overall linear over the same routes, can have inconsistent trajectories, mainly due to fluctuating weather or congestion problems. This results in ADS-B time-series having a tolerance margin when used to train ML models. This means that all attacks made within this margin will likely end up not being detected if the attacker carefully conforms to the ADS-B protocol and to the flight plan. 

\item The DAE in its current state does not support online learning and therefore cannot be updated to the latest ADS-B data. However, all the data used in the evaluation dataset are from 2021 while the training data were from 2020 showing no significant differences between them. This result only has two explanations: either the data does not change significantly enough over time to make a difference or the model is robust enough to not be disturbed by small changes. Only future data will give proper insight to answer this. In addition, the low FPR on the world dataset shows that the model is area-agnostic thanks to the features and the data-processing used for the data. This avoid the training of different models for specific regions.

\item One of the downside of the creation of \emph{realistic} scenarios through a framework like FDI-T is the introduction of a tool bias which could lead to the detection of anomalies being eased. While this would question a supervised approach being trained using said data, for the unsupervised approach, it only shows that models are able to detect these abnormal scenarios. If coupled with a few real life examples of anomaly situation, it only constitutes contents to prove the robustness of the models.

\item The Ryanair anomaly is detected the quickest by the DAE but it is also the model where the anomaly disappear once the main change in track is over. This can be explained by the switch to the DESCENT decoder which is less sensitive to changes in track due to the flight activity when approaching the arrival airport including congestion management and level flight. These results could be improved by adding other decoders taking level phase data or congestion management data.

\item In this experiment, all the decoders of the DAE have the same hyper-parameters and the same architecture. One could decide to make one decoder bulkier or smaller depending on the data fed to it. In the case of the ADS-B, the climbing and the descending data being more complex, it would make sense to have deeper or bigger decoders.

\item Having a FPR higher than zero can be a problem in the air traffic management as it would trigger unnecessary measures to take care of false alarms. Unfortunately, it is not easy task to create a model sensitive enough to detect all kind of anomaly without ever have false positives. On Figure \ref{fig:dae_results}, the false positives observed barely exceed the threshold while the anomaly scores like the one on the Ryanair hijack will almost reach five times the threshold value. Adding other \emph{soft} thresholds -- e.g. four or five sigma rule -- to determine the gravity of the anomaly could help discarding the false alarms in most cases and on the other hand could raise emergencies if the anomaly score would go too high, disregarding entirely the rest of the flight. This strategy would help in the case of anomaly spikes like in the constant position offset.

\end{itemize}

\section{Conclusions and Future Work}
\label{sec:conc}

Detecting anomalies in the ADS-B protocol can greatly improve the monitoring and troubleshooting of the airspace for the air traffic managers in a timely manner. In this paper, we introduced the DAE, a novel auto-encoder architecture to detect anomalies in ADS-B multivariate time-series which work efficiently in any ADS-B covered area, with low chances of false alarms. 
Thanks to a complete data acquisition framework and tools available online, a baseline dataset was created to train, validate and evaluate machine learning models implementation, also fully open-accessible. The results presented on this dataset show that the DAE model perform as well or better than other comparable anomaly detection models and can be more reactive on emergencies. 

Potential future work following these results can be separated in two main areas:

\itemize 

\item First, strong beliefs upon the re-usability of this architecture will be audited in the ATC domain using other discriminating features like the type of aircraft or the kind of sensors used to gather the ADS-B data. Other data could also complement the current ADS-B data to improve the accuracy of anomaly detection models like the DAE. Weather data could be added to better isolate storm avoidance manoeuvres from anomalous situations. The use of the COMM-B data, which are information also broadcast by the aircraft, could also help to determine the authenticity of the ADS-B. The use of these other forms of data will require data collection efforts as they are not as easily accessible as the ADS-B.

\item In order to validate the DAE approach, there is a need to check its efficacy in other domains. The maritime domain uses the AIS protocol, very similar to ADS-B, with equivalent cyber-security issues. Experiments could be done using the DAE with discriminating features like the tonnage of the vessels, its function or its distance from the shore. 

\section{CRediT author statement}
\label{sec:cred}
\textbf{Antoine Chevrot:} Conceptualization, Methodology, Software, Data Curation, Visualization, Writing - Original Draft. \textbf{Alexandre Vernotte:} Investigation, Writing - Review \& Editing. \textbf{Bruno Legeard:} Writing - Review \& Editing, Supervision.

\section{Acknowledgment}
\label{sec:ack}
This work is part of an ongoing research initiative toward the detection of 
FDIA in air traffic surveillance communication flows. This work is partially supported by a UBFC-ISITE-BFC SARCoS Grant, the EIPHI Graduate school (contract \emph{ANR-17-EURE-0002}), and ANR GeLeaD project funding. All Computations have been performed on the supercomputer facilities of the Mésocentre de calcul de Franche-Comté. 

\section{Declaration of Competing Interest}
\label{interest}
The authors declare that they have no known competing financial interests or personal relationships that could have appeared to influence the work reported in this paper.

\bibliographystyle{abbrvnat}
\bibliography{DAE}   

\end{document}